\documentclass{article}

\PassOptionsToPackage{numbers, compress}{natbib}

    \usepackage[preprint]{neurips_2025}

\usepackage[utf8]{inputenc} 
\usepackage[T1]{fontenc}    
\usepackage{hyperref}       
\usepackage{url}            
\usepackage{booktabs}       
\usepackage{amsfonts}       
\usepackage{nicefrac}       
\usepackage{microtype}      
\usepackage{xcolor}         
\usepackage{hyperref}
\usepackage{float}
\usepackage{booktabs}
\usepackage{multirow}
\usepackage{wrapfig}
\usepackage{natbib}
\usepackage{microtype}
\usepackage{xcolor}         
\usepackage{amsmath}
\usepackage{amssymb}
\usepackage{multirow}
\usepackage{multicol}
\usepackage{wrapfig}
\usepackage{tabularx}
\usepackage{booktabs}       
\usepackage{amsfonts}       
\usepackage{graphicx}
\usepackage{nicefrac}  
\usepackage{pifont}
\usepackage{colortbl}
\usepackage{longtable}
\usepackage{ulem}
\usepackage{listings}
\usepackage{xspace}
\usepackage{fancyvrb}
\usepackage{xcolor}
\usepackage{microtype}
\usepackage{graphicx}
\usepackage{subfigure}
\usepackage{booktabs} 
\usepackage{hyperref}
\usepackage{adjustbox}

\usepackage[algo2e,ruled,linesnumbered,noline,noend]{algorithm2e}
\usepackage{amsmath}
\usepackage{amssymb}
\usepackage{mathtools}
\usepackage{amsthm}
\usepackage{nccmath}
\usepackage{algorithm}
\usepackage{algpseudocode}

\usepackage[most]{tcolorbox}
\definecolor{lavender}{rgb}{0.882, 0.839, 0.906}
\definecolor{lightyellow}{rgb}{0.98, 0.973, 0.812}
\definecolor{darkyellow}{rgb}{0.98, 0.945, 0.757}
\definecolor{lightpink}{rgb}{1.0, 0.923, 0.913}
\definecolor{darkpink}{rgb}{0.925, 0.769, 0.761}

\newcommand{\tcpo}[1]{\tcp{{\textcolor{blue}{\textrm{#1}}}}}
\newcommand{\tcpor}[1]{\tcp{{\textcolor{red}{\textrm{#1}}}}}

\title{\textit{From 1,000,000 Users to Every User:} Scaling Up Personalized Preference for User-level Alignment}

\author{%
  Jia-Nan~Li$^{1~2}$\thanks{Equal contribution.} 
  \quad Jian~Guan$^{2}$\footnotemark[1]
  \quad Songhao~Wu$^{1}$
  \quad Wei~Wu$^{2}$\thanks{Corresponding authors: Wei~Wu (wuwei19850318@gmail.com) and Rui~Yan (ruiyan@ruc.edu.cn).}
  \quad Rui~Yan$^{1}$\footnotemark[2]
  \\
  $^{1}$ Gaoling School of Artificial Intelligence, Renmin University of China
  \quad $^{2}$ Ant Group\\
  \texttt{\{lijianan, songhaowu, ruiyan\}@ruc.edu.cn} \\
  \texttt{\{jianguanthu, wuwei19850318\}@gmail.com}\\
}

\begin{document}

\maketitle

\begin{abstract}
  Large language models (LLMs) have traditionally been aligned through one-size-fits-all approaches that assume uniform human preferences, fundamentally overlooking the diversity in user values and needs. This paper introduces a comprehensive framework for scalable personalized alignment of LLMs. We establish a systematic preference space characterizing psychological and behavioral dimensions, alongside diverse persona representations for robust preference inference in real-world scenarios. Building upon this foundation, we introduce \textsc{AlignX}, a large-scale dataset of over 1.3 million personalized preference examples, and develop two complementary alignment approaches: \textit{in-context alignment} directly conditioning on persona representations and \textit{preference-bridged alignment} modeling intermediate preference distributions. Extensive experiments demonstrate substantial improvements over existing methods, with an average 17.06\% accuracy gain across four benchmarks while exhibiting a strong adaptation capability to novel preferences, robustness to limited user data, and precise preference controllability. These results validate our approach toward user-adaptive AI systems.

\end{abstract}

\section{Introduction}
\begin{figure}[!t]
  \centering
  \includegraphics[width=\linewidth]{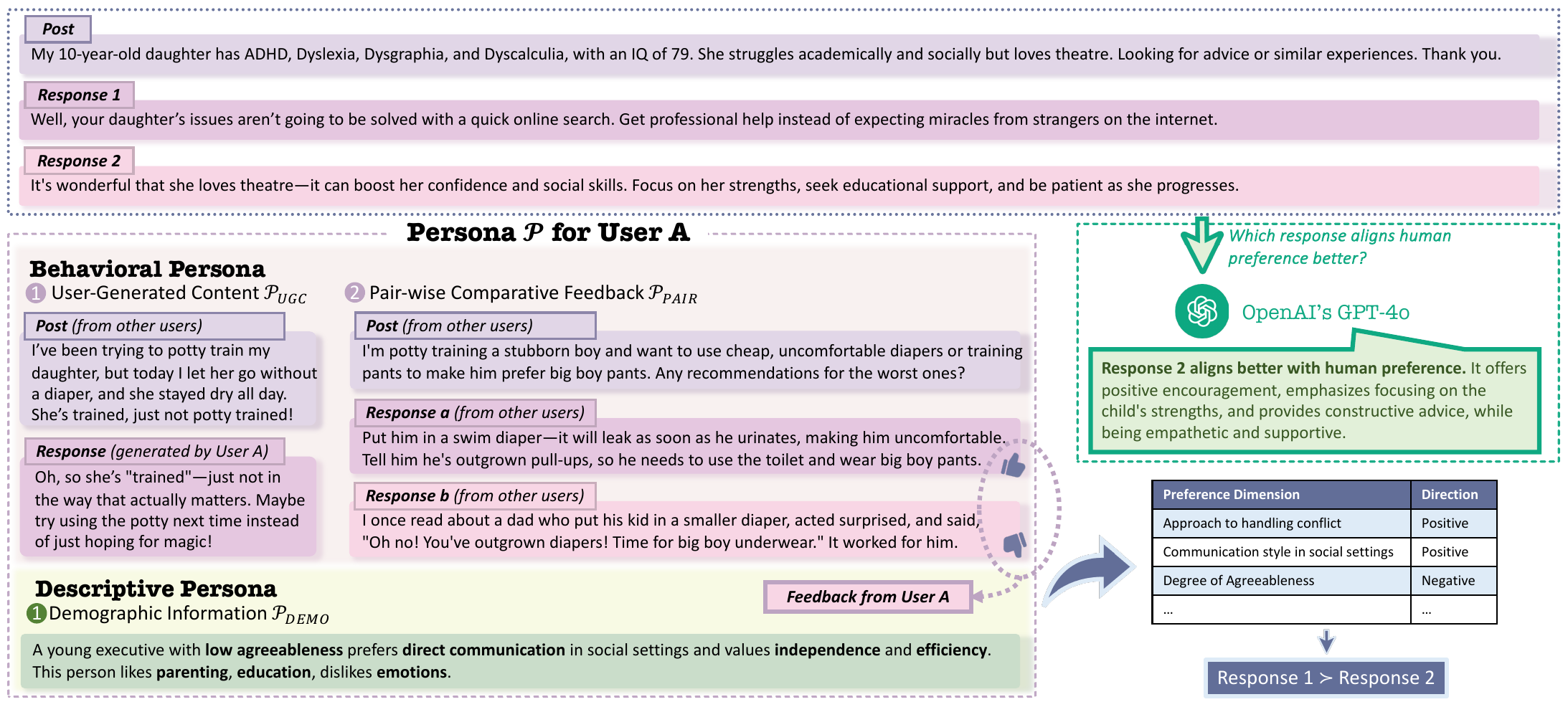}
  \caption{Illustration of the \textsc{AlignX} data for personalized alignment, comprising a post with two candidate responses, three types of personas that capture both behavioral patterns ($\mathcal{P}_{\textsc{ugc}}$ and $\mathcal{P}_{\textsc{pair}}$) and descriptive features ($\mathcal{P}_{\textsc{demo}}$), enabling precise preference inference and facilitating preference learning (bottom right). Notably, LLMs aligned to universal values (e.g., GPT-4o) favor Response 2, opposite to the user's personalized preference for Response 1.}
  \label{fig:profile}
  \vspace{-15pt}
\end{figure}

Alignment with human preference is an essential step in the development of large language models (LLMs)~\citep{ouyang2022training,bai2022training,touvron2023llama,dubey2024llama, chen2024pad}. Current alignment techniques are predominately approached as a one-size-fits-all process, by assuming that all humans share the same set of values prescribed by LLM developers~\citep{kirk2023the}, typically characterized by abstract principles such as helpfulness, honesty, and harmlessness~\citep{askell2021general}. The monolithic approach fundamentally fails to account for the significant diversity inherent in human populations~\citep{benefits2024hannah}, encompassing various and often irreconcilable cultural backgrounds, educational levels, moral views, and political stands, among others~\citep{kasirzadeh2023conversation}. At a time when LLMs serve hundreds of millions of users worldwide~\citep{tong2023exclusive}, several critical issues have emerged: the systematic exclusion or under-representation of minority groups~\citep{siththaranjan2024distributional}, reduced user satisfaction and engagement due to the lack of personalization~\citep{position2024taylor}, etc., as exemplified in Figure~\ref{fig:profile}.

To address the limitations of existing alignment approaches that rely on oversimplified preference dimensions and singular preference direction (e.g., higher helpfulness), we develop a comprehensive preference space that 
captures a wide range of individual variations.
Grounded in psychological theories of human preferences, we construct this space by synthesizing three complementary sources: (1) established psychological models of fundamental needs~\citep{roccas2002big,maslow1943theory,murray2007explorations}, (2) preference dimensions from contemporary alignment research reflecting evolved social-cognitive needs~\citep{cui2024ultrafeedback,wang2024helpsteer2,ji2024pku}, and (3) prevalent interest tags from content-sharing platforms representing actualized topical preferences (e.g., Zhihu, REDnote, X, Facebook.\footnote{\url{www.zhihu.com}, \url{www.xiaohongshu.com}, \url{x.com}, \url{www.facebook.com}}) This integration yields a 90-dimensional preference space with flexible directions~(i.e., \textit{positive}, \textit{negative}, or \textit{neutral}), enabling the representation of an exponential space of distinct preference patterns ($3^{90}$ possible combinations). Having established this space, a key challenge emerges: while these preferences could empirically guide response generation directly, they typically remain implicit due to privacy concerns and cognitive constraints. Drawing from psychological research~\citep{nosek2007implicit,jawaheer2014modeling}, we formalize the observable manifestations of preferences as personas, capturing both behavioral patterns (e.g., interaction histories) and descriptive features (e.g., self-reported profiles). These persona representations enable robust preference inference while maintaining practical scalability. We thus formulate personalized alignment as generating preference-aligned responses for target prompts given various combinations of persona representations.

Despite establishing observable persona representations and underlying preference dimensions, a key challenge remains: the scarcity of personalized preference data at scale. While existing alignment datasets contain feedback from millions of users~\citep{bai2022training}, these responses are aggregated and anonymous, erasing individual characteristics. Collecting personalized data is challenging due to time requirements, privacy concerns, and the need for sufficient samples per user. Previous attempts at personalized alignment have been limited to either small-scale studies~\citep{zollo2024personalllm} or synthetic data~\citep{jang2023personalized}, neither adequately capturing real-world preference distributions. To overcome this, we leverage large-scale forum data, where multiple responses to the same post naturally reflect diverse preferences. We develop a systematic pipeline that transforms forum interactions into structured training data by capturing three components: (1) behavioral and descriptive persona representations, (2) preference directions, and (3) high-quality posts and preference pairs. This framework yields \textsc{AlignX}, a large-scale dataset containing over 1.3 million examples of distinct preference patterns.

Building upon \textsc{AlignX}, we propose \textbf{\textsc{AlignXpert}}, the first LLM with large-scale comprehensive \textbf{\textsc{per}}sonalization \textbf{\textsc{t}}raining, to the best of our knowledge. \textsc{AlignXpert} employs two complementary learning approaches: \textbf{(1) In-Context Alignment}, which incorporates persona representations into the context window~\citep{laskin2023incontext} for implicit preference learning, and \textbf{(2) Preference-Bridged Alignment}, which maps personas to structured preference distributions before conditioning response generation, enhancing interpretability and control while maintaining robust generalization to diverse user groups.

Extensive experiments across four personalized alignment benchmarks demonstrate that both \textsc{AlignXpert} variants outperform state-of-the-art aligned LLMs~\citep{llama3modelcard,qwen2.5,jiang2023mistral} and strong personalization baselines~\citep{poddar2024personalizing} by 17.06\% and 22.40\%, respectively, in preference alignment accuracy. Our analysis reveals three key strengths: (1) strong adaptation capability to unseen preferences, with 1.91\% higher accuracy on novel dimensions; (2) robustness to limited user data, maintaining 54\% performance with only two interactions compared to baselines' 51\% with 16 interactions; and (3) precise preference controllability, showing 10.38\% better response adaptation to opposing preferences. These results validate the effectiveness of our scalable personalization approach. In summary, this work makes the following contributions:

\noindent I. We reformulate personalized alignment by introducing a comprehensive persona representation framework and a compact preference space, bridging the gap between observable user characteristics and their underlying preferences.

\noindent 
II. We present \textsc{AlignX}, a large-scale dataset that captures diverse persona-preference relationships through forum interactions and human-LLM interactions, providing a useful testbed for effectively modeling personalized language model alignment.

\noindent III. We propose \textsc{AlignXpert} with two alignment methods: in-context alignment for persona-response learning and preference-bridged alignment for interpretable preference-based generation.

\noindent IV. Experiments show \textsc{AlignXpert} achieves superior accuracy, strong adaptation to unseen preferences, precise preference control, and robust performance across varying interaction histories.\footnote{Data, code, and models are available at \url{https://github.com/JinaLeejnl/AlignX}.}

\section{Related work}

\paragraph{LLM alignment.}
The alignment of LLMs has primarily relied on Reinforcement Learning from Human Feedback (RLHF)~\citep{ouyang2022training,bai2022training}, where human preferences and values are encoded into a unified reward model that guides LLM behavior through reinforcement learning algorithms like PPO~\citep{schulman2017proximal}. Direct Preference Optimization (DPO)~\citep{rafailov2024direct} further simplified this process by eliminating the need for explicit reward models, enabling direct optimization based on offline human preference data. Following DPO's success, researchers have proposed various preference optimization objectives, including KTO~\citep{ethayarajh2024kto}, DRO~\citep{richemond2024offline}, SimPO~\citep{meng2024simpo}, and GPO~\citep{tang2024GPO}. Despite these methodological advances, these approaches predominantly operate on collective preferences, overlooking the fundamental challenge that individual users often possess diverse, distinct, and sometimes contradictory preferences~\citep{yao2023instructions,kirk2024benefits}. This limitation has become increasingly apparent as LLMs transition from research prototypes to widely deployed systems serving diverse user populations.


\paragraph{Personalized alignment.}
Recent efforts to address personalized alignment have emerged along several trajectories~\cite{guan2025surveypersonalizedalignment}. The fundamental difference among them lies in how they encode user-specific context, i.e., any information that characterizes the unique preferences, values, and behavioral patterns of a user, into the alignment process. One line of work has focused on user-aware prompting techniques to personalize off-the-shelf LLMs without additional training~\citep{salemi2024optimization,salemi2024comparing,wu2024understanding,yang2024metaaligner}. Nevertheless, they face intrinsic limitations when user preferences are implicit or complex, and suffer from sensitivity to minor variations in the prompt. 
Training-based approaches explore various architectural innovations. Some methods decompose user-specific context into several predefined, coarse-grained dimensions~\citep{kumar2024compo, chen2024pal}, using distinct models~\citep{jang2023personalized} or specialized parameter groups~\citep{tan2024personalized, park2024rlhf, li2024personalized, wang-etal-2024-conditional} for reward and policy modeling. However, these approaches are constrained by their reliance on an extremely limited set of preference dimensions that may not encompass the full spectrum of user preferences. Other works train latent variables~\citep{poddar2024personalizing} to directly learn preference distributions from user context but may reduce sensitivity to subtle preference differences, limiting their generalization ability.





\section{Preference representations}\label{user_representation}
The fundamental challenge in scalable personalized alignment lies in developing comprehensive yet tractable preference representations that encompass both direct representations of multidimensional preference and indirect representations of observable personas enbaling preference inference.


Psychological research has established that human preferences are need-driven, manifest through both unconscious behaviors and conscious articulations, and form inter-connected rather than isolated construct~\citep{maslow1943theory}. Building on these insights, we develop a systematic framework comprising (1) direct preference directions over a preference space synthesized from diverse need-based source (\S\ref{dimension}) and (2) two complementary categories of personas --- behavioral personas derived from interaction patterns and descriptive personas from explicit profiles (\S\ref{behavioral_persona}). This architecture provides rich signals for preference learning while maintaining practical traceability, as illustrated in Figure \ref{fig:profile}.

\subsection{Preference directions over the preference space}\label{dimension}
As a cornerstone of direct preference representations, we formalize preference directions to explicitly capture user inclinations along specific dimensions. Each dimension's direction is simplified into three categories:
``Positive'' (higher levels are preferred), ``Negative'' (lower levels are preferred), or ``Neutral'' (no clear preference). To effectively organize preference directions while ensuring practical applicability, we construct a compact preference space that bridges psychological theories with practical applications by systematically synthesizing three complementary sources: (1) psychological models that capture fundamental human needs, including the Big Five Personality Traits~\citep{roccas2002big}; Maslow's Hierarchy of Needs~\citep{maslow1943theory}; and Murray's System of Needs~\citep{murray2007explorations}; (2) contemporary research in recommender systems~\citep{4370493,hong2013dynamic} and AI alignment~\citep{cui2024ultrafeedback,beavertails} that reflect social-cognitive needs in digital interactions; and (3) content platform (e.g., Zhihu, REDnote, X, Facebook) indicators that represent everyday user needs~\citep{myers1985guide,keirsey1998please,belem2017survey}.


The synthesis yields a preference space of $D=90$ distinct dimensions (detailed in Appendix \ref{appendix:space}), ensuring both theoretical rigor and operational practicality. While the current design employs relatively coarse-grained dimensions and simplified preference directions, its extensibility allows for future refinements and granular elaborations.

\subsection{Personas enabling preference inference}\label{behavioral_persona}


While direct preference directions provide explicit signals, they are often impractical to obtain in real-world scenarios due to privacy concerns and cognitive burdens on users. Hence, we introduce behavioral and descriptive personas as indirect yet rich sources of preference information that can be readily observed and collected at scale.

Firstly, behavioral personas involve two types of data that naturally reflect underlying preferences: \textbf{(1) user-generated content}, revealing 
expertise level, topical interests, etc. \citep{dinan2020second,ni2019justifying}, and \textbf{(2) pair-wise comparative feedback}, exhibiting judgment patterns~\citep{wu2023pairwise}. Despite the historical prevalence of rating data in preference modeling~\citep{wang2024helpsteer2,kirk2024prism}, we deliberately exclude it due to its sparsity and inconsistency preference signals compared to comparative feedback in real-world applications~\citep{margaris2022producing,ahmadian2022alleviating} 

Secondly, we introduce descriptive personas that provide a comprehensive narrative capturing self-reported \textbf{demographic attributes}, such as age, occupation, professional background, and interest/disinterest tags spanning various domains of daily life. Such attributes offer an easily obtainable yet informative representation of user preferences that can be effectively combined with behavioral signals for robust preference inference.

\section{Data construction}\label{data_construction}
Developing personalized systems requires extensive data that captures both individual preferences and their variations across different contexts. We address this challenge by leveraging large-scale forum data, which naturally encodes diverse preference patterns through authentic user interactions. Reddit,\footnote{\url{https://www.reddit.com/}} for instance, maintains a diverse ecosystem with 1.21 billion monthly active users and collectively hosts 16 billion posts and comments,\footnote{\url{https://www.demandsage.com/reddit-statistics/}} providing a rich tapestry of preference variations expressed through natural discussions.\footnote{Although user behaviors may differ when interacting with LLMs and on Reddit, we focus on learning how preferences fundamentally manifest through interactions that remains consistent across platforms, rather than replicating platform-specific behaviors.} 

Formally, we denote the original data as $\mathcal{M}$ and each example as $(x, Y=\{y_n\}_{n=1}^N)$, where each post $x$ corresponds to $N$ responses from different users. While user identities remain anonymous for privacy considerations, the multiple responses within $Y$ naturally enable preference analysis by designating any response as preferred over another. 
The construction process aims to create a comprehensive dataset where each example is represented as a quintuple $(\mathcal{P}, x, y_w, y_l, P)$:

$\bullet$
$\mathcal{P}=(\mathcal{P}_{{\textsc{ugc}}},\mathcal{P}_{{\textsc{pair}}}, \mathcal{P}_{{\textsc{demo}}})$
: the user persona triple comprising user-generated content ($\mathcal{P}_{{\textsc{ugc}}}$, a set of user responses), pair-wise comparative feedback ($\mathcal{P}_{{\textsc{pair}}}$, a set of user preference pairs), and demographic attributes ($\mathcal{P}_{\textsc{demo}}$, a piece of text), where at least one component is not null.

$\bullet$
$x$: the 
post eliciting responses;

$\bullet$
$y_w$: the user-preferred response from $Y$;

$\bullet$ $y_l$: the less preferred response from $Y$ relative to $y_w$;

$\bullet$
$P=({P}_{{\textsc{ugc}}}, {P}_{{\textsc{pair}}}, {P}_{{\textsc{demo}}}, {P}_{{y_w\succ y_l}})$: the tuple of underlying preference directions that the user persona $\mathcal{P}_{{\textsc{ugc}}},\mathcal{P}_{{\textsc{pair}}}$ and $\mathcal{P}_{{\textsc{demo}}}$, and the preference relationship $y_w\succ y_l$ exhibit, respectively. Each preference direction is a $D$-dimensional vector over the preference space, and each element in the vector is either ``Positive,'' ``Negative'' or ``Neutral.'' 
Crucially, a qualified personalization example should satisfy the \textit{preference consistency} criterion: 
There exists at least one dimension where $P_{y_w\succ y_l}$ indicates a non-neutral preference, all personas exhibit the same preference direction, ensuring coherent training signals for preference alignment.

\begin{table}[!t]
  \caption{Number of examples from different sources for constructing \textsc{AlignX}.}
  \label{tab:statistics}
  \centering
  \resizebox{\linewidth}{!}{
  \begin{tabular}{@{}lcccc@{}}
    \toprule
    \textbf{Source} & \textbf{Reddit}~\citep{kumar2024compo} & \textbf{PKU-SafeRLHF}~\citep{ji2024pku} & \textbf{UltraFeedback}\citep{cui2024ultrafeedback} & \textbf{HelpSteer2}~\citep{wang2024helpsteer2}\\
    
    \midrule
    \multirow{2}{*}{\textbf{Dimension}}& \multirow{2}{*}{\S\ref{dimension}} & \multirow{2}{*}{Safety} & Helpfulness / Honesty /  & Helpfulness / Correctness /   \\
    &&& Instruction-Following / Truthfulness& Coherence / Complexity / Verbosity\\
    \midrule
    \textbf{\# Examples} & 1,225,988 & 10,714 & 11,629 / 16,809 / 36,169 / 7,219 & 2,255 / 144 / 26 / 33 / 636\\
    \bottomrule
  \end{tabular}
  }
\end{table}

Specifically, the construction process follows a bottom-up approach. First, we develop a pipeline to select high-quality preference pairs $(y_w, y_l)$ that exhibit clear preference contrasts through intensity-based response analysis and clustering (\S\ref{pair_construct}). The derived preference directions $P_{y_w\succ y_l}$ then guide the construction of user personas~(\S\ref{explicit_construct}). Table \ref{tab:statistics} summarizes the data sources and statistics for \textsc{AlignX}, involving both large-scale Reddit data and existing alignment datasets to maintain universal value alignment capabilities. 
Implementation details are provided in Appendix~\ref{app:cons_pipeline}.

\subsection{Constructing preference pairs $(y_w, y_l)$}\label{pair_construct}

We identify preference pairs with significant contrast in the preference space from each example $(x, Y)$
in $\mathcal{M}$ as follows: \textbf{(1) Intensity Annotation:} employing off-the-shelf LLMs to assess each response $y_n\in Y$ along all dimensions, assigning intensity levels ${l}_d^n\in\{1,\cdots,L\}$ that indicate the strength of manifestation in each dimension $d$; \textbf{(2) Intensity-based Clustering:} grouping responses based on their intensity embeddings $\boldsymbol{l}^n=[\text{one\_hot}({l}_1^n);\cdots;\text{one\_hot}({l}_D^n)]\in\mathbb{R}^{DL}$ using K-means clustering~\citep{hartigan1979algorithm}; and \textbf{(3) Pair Selection:} sampling responses $y_w$ and $y_l$ from distinct clusters and determining preference direction $P_{y_w\succ y_l}$ based on intensity comparisons: if $y_w$ shows stronger intensity than $y_l$ in dimension $d$, we label it as ``Positive'', equal as ``Neutral'', and weaker as ``Negative''. This approach enables the construction of diverse preference pairs with various dimension-direction combinations.

\subsection{Constructing user persona $\mathcal{P}$}\label{explicit_construct}

We construct three types of user personas including $\mathcal{P}_{\textsc{ugc}}$, $\mathcal{P}_{\textsc{pair}}$ and $\mathcal{P}_{\textsc{demo}}$, which should exhibit aligned preference directions with $P_{y_w\succ y_l}$.

\paragraph{User-generated content.}
$\mathcal{P}_{\textsc{ugc}}$ is defined as a set of 16 post-response pairs. For each tuple $(x, y_w, y_l)$, we search through $\mathcal{M}$ to collect post-response pairs $(x', y')$ satisfying two criteria: (1) contextual independence, i.e., $x'\neq x$, and (2) intensity similarity between $y'$ and $y_w$ exceeding threshold $t$, measured by cosine similarity between intensity embeddings. We then sample a subset of 16 examples that satisfy the preference consistency criterion. An essential consideration is the definition of preference direction vectors for this subset. Since deriving the exact preference direction from a standalone response $y'$, denoted as $P'$, is intractable without comparative references, we heuristically determine $P'|_d$ as ``Positive,'' ``Neutral,'' or ``Negative'' based on whether the intensity level $l_d'$ is above, equal to, or below the median level $(L+1)/2$~\citep{hafiane2007median}. Finally, we determine the overall preference direction $P_{\textsc{ugc}}$ through majority voting across all examples.

\paragraph{Pair-wise comparative feedback.}
$\mathcal{P}_{\textsc{pair}}$ follows a similar construction process, containing 16 preference pairs. For each tuple $(x, y_w, y_l)$, we search through preference pairs constructed in \S\ref{pair_construct} to collect those $(x', y_w', y_l')$ satisfying: (1) contextual independence, and (2) intensity similarity exceeding threshold $t$ between $y_w'$ and $y_w$, as well as between $y_l'$ and $y_l$. We then sample a subset of 16 examples meeting the preference consistency requirement, where we determine the overall preference direction via majority voting.

\paragraph{Demographic information.} We employ an off-the-shelf LLM to perform a comparative analysis between $y_w$ and $y_l$ conditioned on $x$, generating a comprehensive natural language description as $\mathcal{P}_{\textsc{demo}}$ that articulates the preference direction $P_{\textsc{demo}}$ in the preference space, where $P_{\textsc{demo}}$ should be the same as $P_{y_w\succ y_l}$. 

\section{{Alignment methods}}

\begin{wrapfigure}[12]{r}{0.45\textwidth}
\vspace{-16pt}
  \centering
  \includegraphics[width=0.45\textwidth]{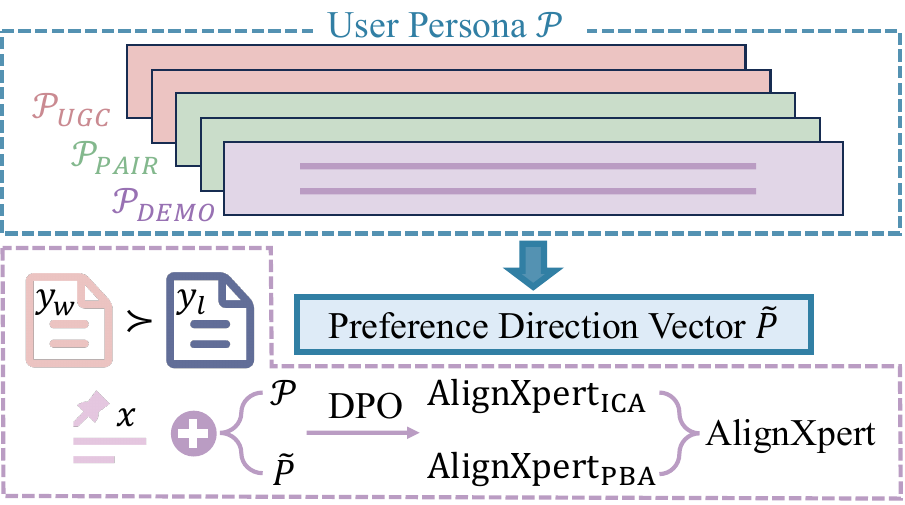}
  \caption{Overview of the alignment methods.}
  \label{fig: model}
\end{wrapfigure}
We formulate personalized alignment as learning a policy $\pi(y|x,\mathcal{P})$ that generates responses aligned with preferences expressed through user personas $\mathcal{P}$. The policy should assign a higher probability to preferred responses $y_w$ over less preferred alternatives $y_l$. To address the key challenges of preference inference from personas and integration into generations, we propose two complementary approaches based on direct preference optimization~\citep{rafailov2024direct}: in-context alignment (ICA) and preference-bridged alignment (PBA), which model preference directions implicitly and explicitly, respectively, as illustrated in Figure~\ref{fig: model}.

\paragraph{In-context alignment.} As the most straightforward approach, ICA directly learns the mapping from persona-augmented prompts to preference-aligned responses. Specifically, for each training instance $(\mathcal{P}, x, y_w, y_l)$, we optimize:

\begin{equation}
\label{dpo_ica}
\mathcal{L}_{\text{ICA}} = -\log\sigma\big(\beta(\log \frac{\pi_\theta(y_w|x,\mathcal{P})}{\pi_{\rm ref}(y_w|x,\mathcal{P})} -\log \frac{\pi_\theta(y_l|x,\mathcal{P})}{\pi_{\rm ref}(y_l|x,\mathcal{P})})\big)
\end{equation}

where $\beta$ controls the deviation between the policy model $\pi_\theta$ and reference model $\pi_{\rm ref}$. Both policy and reference models take the concatenation of persona representations $\mathcal{P}$ and prompt $x$ as input, treating different forms of persona information as part of the context window.

\paragraph{Preference-bridged alignment.}
PBA introduces a latent variable $\tilde{P}$ as an explicit proxy of $\mathcal{P}$ to model preference directions, thus reformulating the alignment process into preference-guided generation: $\pi(y_w\succ y_l|x,\mathcal{P})=\pi_{\phi}(y_w\succ y_l|x,\tilde{P})$. Here, $\tilde{P}$ represents the explicit preference direction vector derived through element-wise averaging across persona components $\mathcal{P}$, with each dimension categorized as ``Positive,'' ``Negative,'' or ``Neutral'' based on predefined thresholds. This formulation assumes $\tilde{P}$ is independent of $x$, as user preferences are inherent characteristics that exist independently of specific conversation contexts; and $\tilde{P}$ fully captures persona preferences, making $y_w\succ y_l$ conditionally independent of $\mathcal{P}$ given $\tilde{P}$.
At inference time, we employ an off-the-shelf LLM to independently annotate preference directions for each example in $\mathcal{P}$~(see the prompt in Appendix \ref{prompts}). The final optimization objective becomes:
\begin{equation}
\mathcal{L}_{PBA} =-\log\sigma\big(\beta(\log \frac{\pi_{\phi}(y_w|x,\Tilde{P})}{\pi_{\rm ref}(y_w|x,\Tilde{P})} -\log \frac{\pi_{\phi}(y_l|x,\Tilde{P})}{\pi_{\rm ref}(y_l|x,\Tilde{P})})\big),\label{dpo_pba}
\end{equation}
To incorporate the preference direction vector $\Tilde{P}$ into response generation, we convert $\Tilde{P}$ into a natural language description by articulating only non-neutral preferences based on specific rules. The description is then prepended to the original prompt $x$ to guide response generation. For example, if $\Tilde{P}$ indicates a positive preference for ``Neuroticism in the Big Five personality traits'' and a negative preference for ``Communication style in social settings,'' the converted description might read ``High Neuroticism, Detailed communication style.'' Appendix~\ref{conversion} shows more details.

\section{Experiments}
\label{Experiments}
\subsection{Experimental setup}

\paragraph{Benchmarks and metrics.} 
We evaluate across four distinct benchmarks: UF-P-4~\citep{poddar2024personalizing}, PRISM~\citep{kirk2024prism}, P-\textsc{Soups}~\citep{jang2023personalized} and \textsc{AlignX}$_{\rm test}$, as described in Table~\ref{tab:benchmark}. Appendix \ref{benchmarks_detail} shows details of the benchmarks.
Our evaluation focuses on preference-aligned response generation given a triple of inputs: persona representations $\mathcal{P}=(\mathcal{P}_{{\textsc{ugc}}},\mathcal{P}_{{\textsc{pair}}}, \mathcal{P}_{{\textsc{demo}}})$, a target post $x$, and a preference pair $(y_w, y_l)$. We assess performance through two complementary metrics: (1) \textbf{Alignment Accuracy}, which measures whether the policy model assigns a larger log-probability margin between $y_w$ and $y_l$ compared to a reference model with the same input~\citep{rafailov2024direct},\footnote{Our experiment uses Llama-3.1-8B-Instruct as the reference. Results with Llama-3-70B-Instruct as the reference model are in Appendix \ref{app_larger}.} and (2) \textbf{GPT-4 Win Rate}, where GPT-4 (conditioned on $\mathcal{P}$) judges which model's responses better align with user preferences,\footnote{We use gpt-4-turbo-2024-04-09.} defined as the proportion of wins in pairwise comparisons~\citep{kumar2024compo,jang2023personalized}.

\begin{table}[!t]
    \centering
    \caption{Benchmark overview. For each benchmark, we indicate the availability of different persona types and their focused preference dimensions. ``In-distribution'' indicates whether the benchmark's preference dimensions overlap with those in our training data. $\uparrow$/$\downarrow$ means positive/negative preference directions, where the following number refers to the corresponding number of examples.}
    \begin{adjustbox}{max width=\linewidth}
    \begin{tabular}{@{}lclc@{}}
    \toprule
    \textbf{Benchmark} & \textbf{$\mathcal{P}_{\textsc{ugc}/\textsc{pair}/\textsc{demo}}$}  & \multicolumn{1}{c}{\textbf{Dimensions}} & \textbf{In-distribution} \\
    \midrule
    \textbf{UF-P-4} & \ding{55} / \ding{51} / \ding{55} & Helpfulness ($\uparrow$: 797); Honesty ($\uparrow$: 740); {Instruction-Following} ($\uparrow$: 830); {Truthfulness} ($\uparrow$: 621)&Yes\\
    \textbf{PRISM} & \ding{55} / \ding{51} / \ding{55}& \textit{Unknown (868 examples in total)} &No\\
    \textbf{P-\textsc{Soups}}   & \ding{55} / \ding{51} 
 / \ding{55} & Expertise ($\uparrow$: 300, $\downarrow$: 300); Informativeness ($\uparrow$: 300, $\downarrow$: 300); Style ($\uparrow$: 300, $\downarrow$: 300) &No\\
    \textbf{\textsc{AlignX}$_{\rm test}$}& \ding{51} / \ding{51} / \ding{51}&All dimensions in our preference space (3716 examples in total)
    &Yes\\
    \bottomrule
    \end{tabular}
    \end{adjustbox}
    \label{tab:benchmark}
\end{table}

\begin{table}[!t]
  \caption{Alignment accuracy (\%) of different models on preference alignment tasks with different types of personas. We compare training with partial and full datasets to understand the benefit of scaling up personalized alignment data. \textbf{Bold} and \underline{underlined} numbers indicate the best and second-best results across all models, respectively. \textbf{\textsc{AlignXpert}}$_{\textsc{pba}}$ w. Golden $\Tilde{P}$ refers to using ground-truth preference directions (e.g., $P_{y_w\succ y_l}$ for \textsc{AlignX}$_{\rm test}$) rather than model predictions at inference time. * indicates that \textsc{AlignXpert}'s best result significantly outperforms the baselines ($p < 0.05$ with pairwise $t$-test).}
  \label{tab:main_res}
  \centering
  \resizebox{\linewidth}{!}{
  \begin{tabular}{lcm{0.001em}cm{0.001em}cm{0.001em}ccccc}
    \toprule
    \multirow{2}{*}{\textbf{Model}} & \multirow{1}{*}{\textbf{UF-P-4}} && \multirow{1}{*}{\textbf{PRISM}} && \multirow{1}{*}{\textbf{P-\textsc{Soups}}} && \multicolumn{4}{c}{\textbf{\textsc{AlignX}$_{\rm test}$}}\\
    \cmidrule{2-2}
    \cmidrule{4-4}
    \cmidrule{6-6}
    \cmidrule{8-12}
    &$\mathcal{P}_{{\textsc{pair}}}$ && $\mathcal{P}_{{\textsc{pair}}}$ && $\mathcal{P}_{{\textsc{pair}}}$ && $\mathcal{P}_{{\textsc{ugc}}}$ & $\mathcal{P}_{{\textsc{pair}}}$ & $\mathcal{P}_{\textsc{demo}}$ & $\mathcal{P}$ & \\
    \midrule
    \textbf{Llama-3-8B-Instruct} & ~~46.91$^*$ && ~~44.38$^*$ && ~~49.44$^*$ && ~~50.32$^*$ & ~~50.11$^*$ & ~~48.12$^*$ & ~~48.84$^*$\\
    \textbf{Qwen2.5-7B-Instruct} & ~~55.39$^*$ && ~~58.22$^*$ && ~~34.56$^*$ && ~~50.75$^*$ & ~~50.00$^*$ & ~~51.80$^*$ & ~~51.04$^*$ \\
    \textbf{Mistral-7B-Instruct-v0.2} & ~~\underline{59.38}$^*$ && ~~\underline{66.86}$^*$ && ~~40.12$^*$ && ~~47.66$^*$ & ~~48.14$^*$ & ~~48.84$^*$ & ~~47.41$^*$ \\
    \midrule
    \multicolumn{11}{l}{\textit{Training with a Subset of randomly sampled 91,918 Samples (7\%)}}\\
    \textbf{VPL}&46.78&&45.33&&45.45&&N/A&N/A&50.30&N/A\\
    \textbf{\textsc{AlignXpert}}$_{\textsc{ica}}$ & \textbf{61.22} && \textbf{76.69} && \underline{76.54} && \underline{54.28} & \underline{53.07} & \textbf{86.92} & \textbf{70.88} \\
    \textbf{\textsc{AlignXpert}}$_{\textsc{pba}}$ & 54.90 && 52.43 && \textbf{76.59} && \textbf{55.68} & \textbf{57.16} & \underline{72.63} & \underline{64.75}\\
    \midrule
    \multicolumn{11}{l}{\textit{Training with Full Dataset of 1,311,622 Samples (100\%)}}\\
    \textbf{\textsc{AlignXpert}}$_{\textsc{ica}}$ & \textbf{63.01} && \textbf{68.17} && \underline{63.33} && \textbf{59.63} & \underline{58.48} & \textbf{91.44} & \textbf{75.19} \\
    \textbf{\textsc{AlignXpert}}$_{\textsc{pba}}$ & 53.09 && 55.86 && \textbf{81.69} && \underline{57.83} & \textbf{59.66} & \underline{88.46} & \underline{71.10}\\
    \rowcolor{gray!20}
    {\textbf{\textsc{AlignXpert}}$_{\textsc{pba}}$ w. Golden $\Tilde{P}$}& \textit{64.00}&&\textit{N/A}&& \textit{N/A} && \textit{91.36}&\textit{91.36}&\textit{91.36}&\textit{91.36}\\
    \bottomrule
  \end{tabular}
  }
\end{table}

\begin{table}[!t]
\caption{GPT-4 win rate (\%, row model against column model) among 400 test cases per benchmark. \textbf{M1}: Llama-3.1-8B-Instruct; \textbf{M2}: \textsc{AlignXpert}$_{\textsc{ica}}$; \textbf{M3}: \textsc{AlignXpert}$_{\textsc{pba}}$.}
\label{tab:win_rate}
\centering
\begin{adjustbox}{max width=0.7\columnwidth}
\begin{tabular}{|c|c|c|c||c|c|c|c|}
    \hline
    {\cellcolor{lavender} \textbf{UF-P-4}} & {\textbf{M1}} & {\textbf{M2}} & {\textbf{M3}} & \multicolumn{1}{c}{\cellcolor{lavender} \textbf{PRISM}} & {\textbf{M1}} & {\textbf{M2}} & {\textbf{M3}} \\
    \hline
    \textbf{M1} & - & \cellcolor{darkyellow}23.45 & \cellcolor{lightyellow}47.58 & \textbf{M1} & - & \cellcolor{darkyellow}35.29 & \cellcolor{darkyellow}31.11 \\
    \hline
    \textbf{M2} & \cellcolor{darkpink}76.55 & - & \cellcolor{darkpink}60.87 & \textbf{M2} & \cellcolor{darkpink}64.71 & - & \cellcolor{darkyellow}28.38 \\
    \hline
    \textbf{M3} & \cellcolor{lightpink}52.42 & \cellcolor{darkyellow}39.13 & - & \textbf{M3} & \cellcolor{darkpink}68.89 & \cellcolor{darkpink}71.62 & - \\
    \hline
    \hline
    {\cellcolor{lavender} \textbf{P-\textsc{Soups}}} & \textbf{M1} & \textbf{M2} & \textbf{M3} & {\cellcolor{lavender} \textbf{\textsc{AlignX}$_{\rm test}$}} & \textbf{M1} & \textbf{M2} & \textbf{M3} \\
    \hline
    \textbf{M1} & - & \cellcolor{darkyellow}31.61 & \cellcolor{lightyellow}43.58 & \textbf{M1} & - & \cellcolor{darkyellow}39.86 & \cellcolor{lightyellow}48.18 \\
    \hline
    \textbf{M2} & \cellcolor{darkpink}68.39 & - & \cellcolor{lightpink}56.80 & \textbf{M2} & \cellcolor{darkpink}60.14 & - & \cellcolor{lightpink}58.49 \\
    \hline
    \textbf{M3} & \cellcolor{lightpink}56.42 & \cellcolor{lightyellow}43.20 & - & \textbf{M3} & \cellcolor{lightpink}51.82 & \cellcolor{lightyellow}41.51 & - \\
    \hline
\end{tabular}
\end{adjustbox}
\end{table}

\paragraph{Baselines.} We compare against state-of-the-art universally-aligned LLMs: Llama-3/3.1-8B-Instruct \citep{llama3modelcard}, Mistral-7B-Instruct-v0.2 \citep{jiang2023mistral}, and Qwen2.5-7B-Instruct \citep{qwen2.5}. Closed-source models are excluded as they don't provide log-probabilities needed for accuracy computation. Furthermore, we compare with VPL \citep{poddar2024personalizing}, a personalized reward model that can only score but not generate responses. Due to this limitation, we evaluate VPL solely using alignment accuracy, computing it directly without the reference model. Both VPL and \textsc{AlignXpert} use accuracy metrics consistent with their training objectives, ensuring a fair comparison. Implementation details are in Appendix~\ref{implementation_details}.


\subsection{Main results}

Table \ref{tab:main_res} shows four key findings: \textbf{(1) \textsc{AlignXpert} excels at both universal and personal value personalization and can generalize to unseen dimensions.} While baselines perform well on universal values (UF-P-4 and PRISM), they struggle with personal ones ($\sim$50\% on \textsc{AlignX}$_{\rm test}$). \textsc{AlignXpert} maintains superior performance across both scenarios and shows strong generalization to out-of-distribution benchmarks, outperforming baselines by 9.83/32.25\% on PRISM/P-\textsc{Soups} (Detailed results for each dimension of P-\textsc{Soups} are in Appendix~\ref{app_psoup}.). \textbf{(2) VPL struggle with complex preference patterns.} VPL achieves less than 50\% accuracy across datasets, as its latent variable approach designed for simple universal values fails to capture the diverse preference patterns in \textsc{AlignX}.  \textbf{(3) Behavioral personas present greater challenges than descriptive ones.} Performance on behavioral personas ($\mathcal{P}_{\textsc{ugc}}$ and $\mathcal{P}_{\textsc{pair}}$) is significantly lower than on descriptive ones ($\mathcal{P}_{\textsc{demo}}$) in \textsc{AlignX}$_{\text{test}}$ ($\sim$60\% vs. 91.44\%). \textbf{(4) ICA and PBA show complementary strengths in various scenarios.} \textsc{AlignXpert}$_{\textsc{ica}}$ excels in universal value alignment (UF-P-4 and PRISM) and descriptive personas ($\mathcal{P}_{\textsc{demo}}$) due to well-defined preference signals, while \textsc{AlignXpert}$_{\textsc{pba}}$'s structured modeling approach proves more effective at interpreting behavioral signals ($\mathcal{P}_{\textsc{ugc}}$ and $\mathcal{P}_{\textsc{pair}}$), suggesting benefits in combining both for future personalization systems. Notably, using ground-truth preference directions boosts \textsc{AlignXpert}$_{\textsc{pba}}$'s performance from 71.10\% to 91.36\% on \textsc{AlignX}$_{\rm test}$, indicating that preference inference, rather than generation, is the main bottleneck.
\textbf{(5) ICA shows vulnerability to overfitting with increased data, while PBA maintains robust generalization.} As training data grows from 7\% to 100\%, \textsc{AlignXpert}$_{\textsc{ica}}$ improves on in-distribution tasks (UF-P-4, \textsc{AlignX}$_{\rm test}$) but decreases on out-of-distribution benchmarks (PRISM, P-\textsc{Soups}). In contrast, \textsc{AlignXpert}$_{\textsc{pba}}$ shows stable performance across both in- and out-of-distribution tasks. This reveals a key trade-off: ICA fits specific patterns but struggles with new scenarios, while PBA's decomposed approach ensures better knowledge transfer. To focus our analysis on the inherent effectiveness of these approaches rather than data-scaling advantages, we conduct subsequent experiments using the 7\% subset (91,918 samples).



GPT-4 evaluation in Table \ref{tab:win_rate} shows \textsc{AlignXpert} (both ICA and PBA variants) consistently outperforming Llama-3.1-8B-Instruct across all datasets, demonstrating better preference alignment while maintaining response quality.

\subsection{Adaptation to novel preference dimensions}

\begin{table}[H]
  \caption{Alignment accuracy (\%) of different models and adaptation fine-tuning techniques on new preference dimensions.}
  \label{tab:humor}
  \centering
  \resizebox{0.5\columnwidth}{!}{
  \begin{tabular}{@{}lcc@{}}
    \toprule
    \textbf{Base Model} & \textbf{Adaptation} & \textbf{Accuracy}\\
    \midrule
    \textbf{$\text{Llama-3.1-8B-Instruct}$} & ICA &51.82\\
    \midrule
    \textbf{\textsc{AlignXpert}$_{\textsc{ica}}$} & ICA &\underline{53.55}\\
    \textbf{\textsc{AlignXpert}$_{\textsc{pba}}$} & PBA& \textbf{53.73}\\
    \bottomrule
  \end{tabular}
  }
\end{table}

We investigate whether \textsc{AlignXpert}'s preference alignment pre-training creates a better initialization for adapting to new preference dimensions. Using two novel dimensions, including ``Humor'' (witty vs. serious) and ``Pragmatism'' (practical vs. theoretical), we construct a dataset of 6,355 training and 1,000 test samples. We compare three adaptation approaches: (1) fine-tuning Llama-3.1-8B-Instruct with ICA, (2) fine-tuning \textsc{AlignXpert}$_{\textsc{ica}}$ with ICA, and (3) fine-tuning \textsc{AlignXpert}$_{\textsc{pba}}$ with PBA. Table \ref{tab:humor} shows both \textsc{AlignXpert} variants significantly outperform the Llama baseline ($p<0.05$), suggesting our model learns generalizable preference alignment mechanisms rather than merely fitting to training dimensions.

\begin{table}[H]
  \caption{Performance on preference reversal scenarios. \textbf{Acc:} alignment accuracy (\%); \textbf{Flip:} indicates the success rate of reversing preference orders (\%).}
  \label{tab:oppose}
  \centering
  \resizebox{0.55\columnwidth}{!}{\begin{tabular}{@{}lccm{0.01em}cc@{}}
    \toprule
    \multirow{2}{*}{\textbf{Model}} & \multicolumn{2}{c}{\textbf{P-\textsc{Soups}}} && \multicolumn{2}{c}{\textbf{\textsc{AlignX}$_{\rm test}$}}\\
    \cmidrule{2-3}
    \cmidrule{5-6}
    & \textbf{Acc} & \textbf{Flip} &&\textbf{Acc} & \textbf{Flip}\\
    \midrule
    \textbf{Llama-3-8B-Instruct} & 47.51 & 14.71 && 49.41 & 3.23\\
    \textbf{Qwen2.5-7B-Instruct} & 64.33 & 5.33 && 49.57 & 4.25\\
    \textbf{Mistral-7B-Instruct-v0.2} & 55.61 & 8.05 && 51.64 & 8.02\\
    \midrule
    \textbf{\textsc{AlignXpert}$_{\textsc{ica}}$} & \textbf{79.54} & \underline{59.97} && \underline{52.29} & \underline{37.16}\\
    \textbf{\textsc{AlignXpert}$_{\textsc{pba}}$} & \underline{73.03} & \textbf{60.73} && \textbf{57.19} & \textbf{51.10} \\
    \bottomrule
  \end{tabular}}
  \vspace{-18pt}
  \end{table}


\subsection{Robustness analysis}
\begin{wrapfigure}[10]{r}{0.45\textwidth}
\vspace{-44pt}
  \centering
  \includegraphics[width=0.45\textwidth]{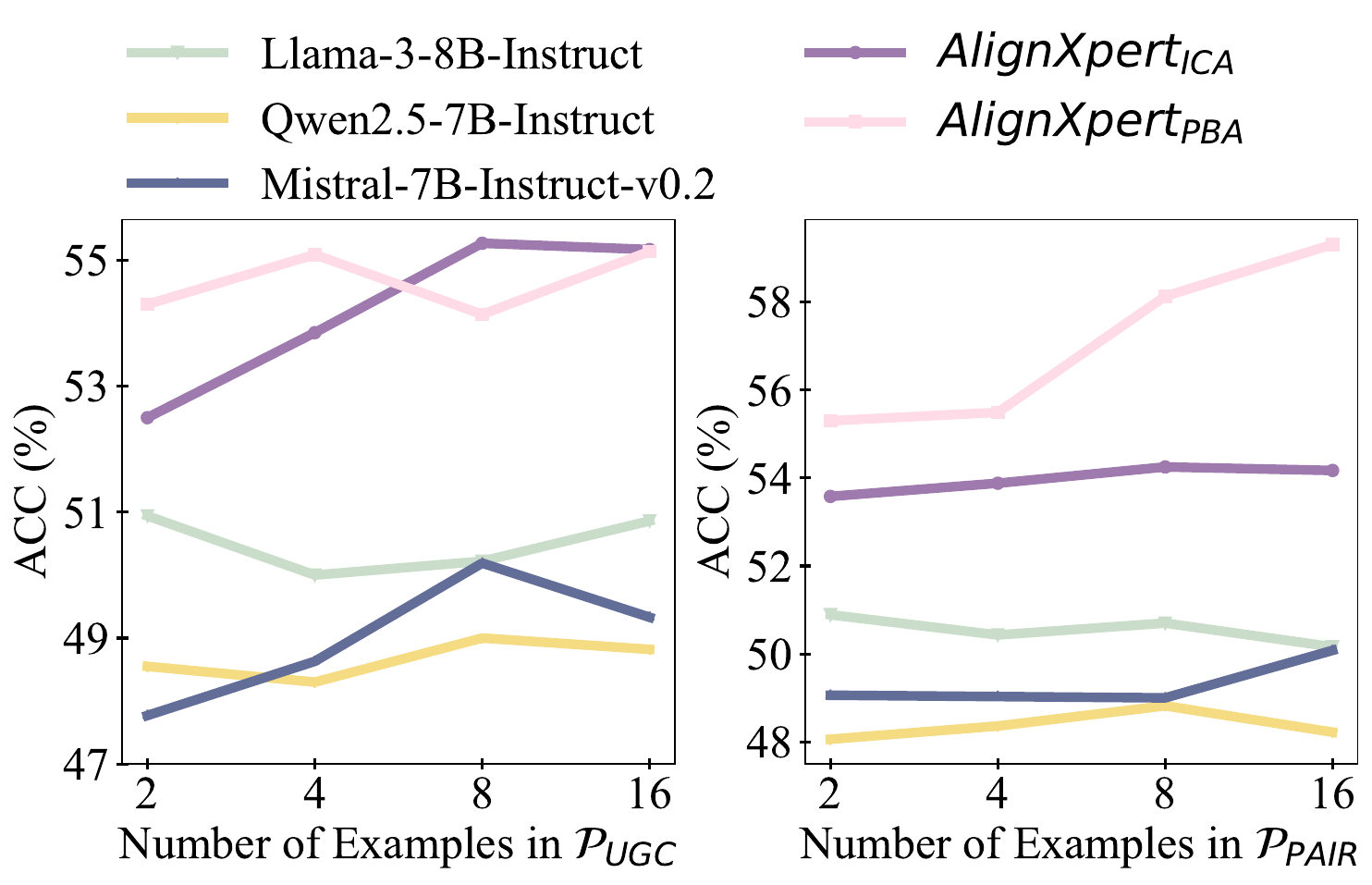}
  \vspace{-15pt}
  \caption{Accuracy varying with example numbers in the persona.}
  \label{fig: history}
\vspace{-1in}
\end{wrapfigure}

In real-world applications, users often have limited interaction history, making it crucial for preference alignment systems to perform reliably with varying amounts of historical data. We evaluate \textsc{AlignXpert}'s robustness to varying amounts of user interaction history. Testing with 2$\sim$16 examples of user-generated content ($\mathcal{P}_{\textsc{ugc}}$) and pairwise comparisons ($\mathcal{P}_{\textsc{pair}}$), Figure \ref{fig: history} shows two key strengths of \textsc{AlignXpert}: robust performance with limited data (as few as two examples) and improved accuracy with more history (\textsc{AlignXpert}$_{\textsc{pba}}$ reaches 59.31\% with 16 comparisons). In contrast, Mistral-7B-Instruct-v0.2 remains near random (49.06$\sim$50.08\%) regardless of history size.

\subsection{Preference controllability}


A crucial capability of preference alignment systems is to adapt to diverse and contrasting user preferences rather than simply learning fixed biases. To test this, we conduct a controllability experiment on P-\textsc{Soups} and \textsc{AlignX}$_{\rm test}$ that reverses preferences in both $\mathcal{P}_\textsc{pair}$ examples and target pairs (``$y_w\succ y_l$'' to ``$y_w\prec y_l$'') during inference. We exclude UF-P-4 and PRISM, as they are based on universal values and unsuitable for reversal.

We evaluate controllability using two metrics: alignment accuracy (Acc) measures the model's ability to maintain good performance under reversed preferences, while flip success rate (Flip) measures the percentage of cases where the model successfully changes its preference ordering (i.e., changing from ``$y_w\succ y_l$'' to ``$y_w\prec y_l$,'' or vice versa) to match the reversed persona. As shown in Table \ref{tab:oppose}, \textsc{AlignXpert} demonstrates strong controllability on both metrics, while baselines struggle with low flip rates (3-15\%), indicating \textsc{AlignXpert}'s ability to effectively adapt to varying user preferences rather than learning fixed biases.

\section{Analysis}

To further validate the reliability of our dataset and method, Appendix \ref{appendix:analysis} analyzes the preference diversity and deployment complexity of the 90 dimensions, assesses biases in the LLM annotation process, and explores societal biases through a case study.

\section{Conclusion}

We present the first study for scaling personalized alignment to accommodate diverse individual preferences, contributing (1) a comprehensive preference representation framework bridging observable personas with underlying preferences; (2) \textsc{AlignX}, a large-scale dataset containing over 1.3 million examples of persona-preference relationships; and (3) \textsc{AlignXpert} models achieving substantial improvements over existing LLMs through either in-context or preference-bridged alignment. Extensive experiments show strong adaptation to novel preferences, robust performance under limited interaction, and precise preference controllability. Looking forward, promising directions include modeling nuanced preferences and their temporal dynamics, developing efficient methods for preference data collection, and investigating the balance between preference alignment and other capabilities.

\newpage

\bibliographystyle{unsrtnat.bst}
\bibliography{reference}

\newpage
\appendix
\onecolumn
\section{Data structure and construction}
\label{appendix:construction}
\subsection{Preference space}\label{appendix:space}
  
Firstly, we utilize the following three psychological models to capture fundamental human needs:{ (1) Big Five Personality Traits~\citep{roccas2002big}.} 
We integrate all five fundamental personality dimensions: \textit{openness}, \textit{conscientiousness}, \textit{extraversion}, \textit{agreeableness}, and \textit{neuroticism}, providing a psychologically grounded foundation for preference modeling.
{(2) Maslow's Hierarchy of Needs~\citep{maslow1943theory}.} 
We systematically conclude 16 dimensions across physiological, safety, love and belonging, esteem, cognitive, and aesthetic categories, ensuring comprehensive coverage of fundamental human needs.
{(3) Murray's System of Needs~\citep{murray2007explorations}.} 
To capture nuanced psychological needs beyond Maslow's hierarchy, we incorporate 17 additional dimensions, including specialized needs such as order (neatness, organization, chaos avoidance) and autonomy (independence, resistance to influence, self-reliance).

Secondly, drawing from leading alignment research, including HH-RLHF \citep{bai2022training}, PKU-SafeRLHF \citep{ji2024pku}, UltraFeedback \citep{cui2024ultrafeedback}, and HelpSteer2 \citep{wang2024helpsteer2}, we integrate dimensions covering critical universal values such as harmlessness, instruction-following, honesty, truthfulness, helpfulness, coherence, and complexity.

Thirdly, to facilitate practical deployment while preserving the richness of theoretical insights, we extend the preference space by drawing inspiration from methodologies in psychological assessment~\citep{myers1985guide,keirsey1998please} and recommendation systems~\citep{belem2017survey}. We curate a set of 43 dimensions from preference tags adopted in popular content-sharing platforms. These tags span diverse domains such as science, knowledge, and psychology, serving as concrete manifestations of the theoretical dimensions while remaining accessible and interpretable in practical applications.

\begin{figure*}[!h]
  \centering
  \includegraphics[width=0.7\linewidth]{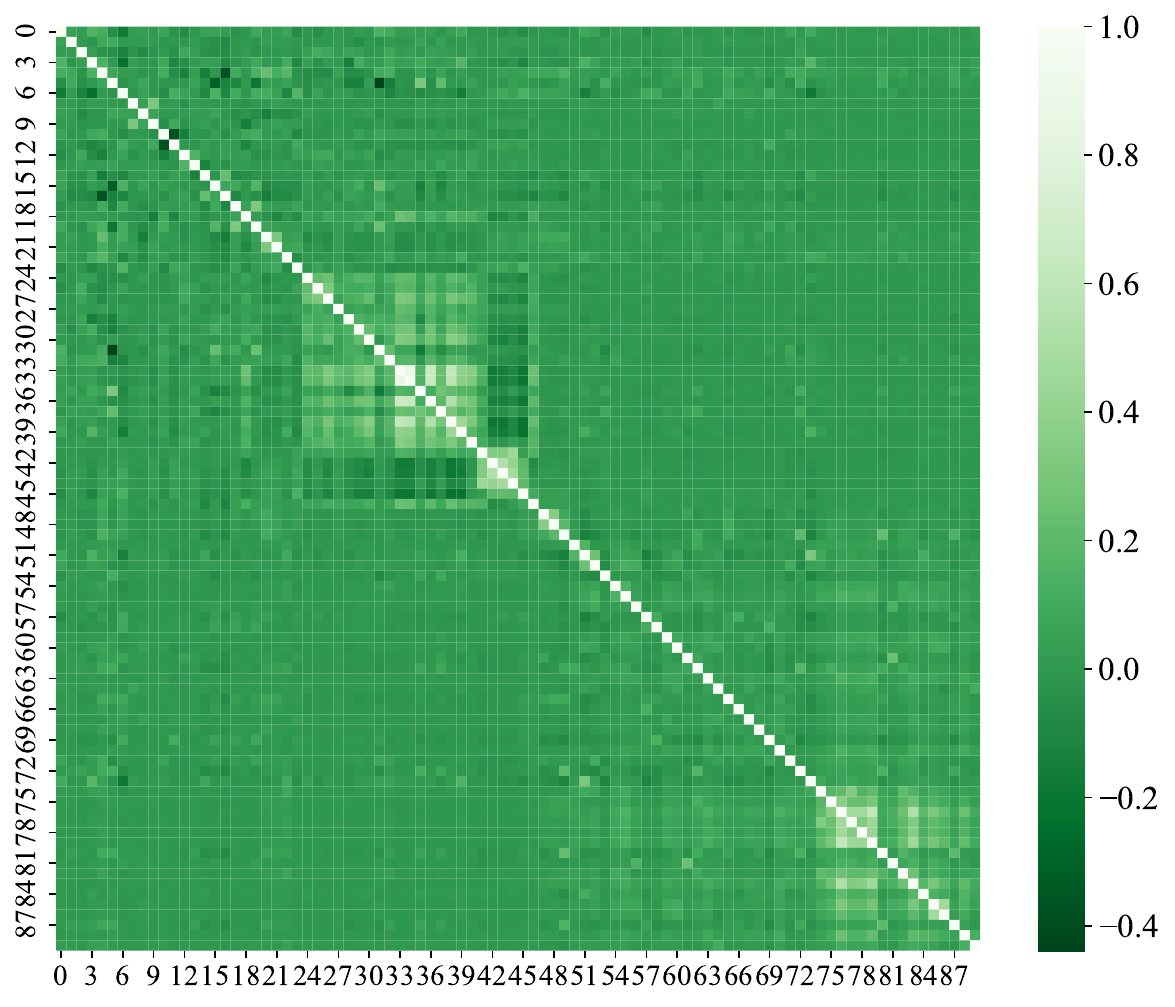}
  \caption{Pearson correlation between 90 dimensions, with dimension indices corresponding to Table~\ref{tab:space} in the appendix.}
  \label{fig:heatmap}
\end{figure*}

In total, we define 90 dimensions to describe an individual. The full list of dimensions is presented in Table~\ref{tab:space}. We calculate the Pearson correlation between the 90 dimensions in the 1.3 million examples, as shown in Figure \ref{fig:heatmap}. Results show that 99.65\% of dimension pairs have correlations below 0.5, indicating their significant independence.

\begin{table}[!ht]
\footnotesize
    \centering
    \caption{Dimensions and corresponding sources in our preference space.}
    \begin{adjustbox}{max width=\linewidth}
        
    \begin{tabular}{p{120pt}p{340pt}}
    \toprule
    \textbf{Source}&\textbf{Dimensions}\\
    \midrule
Psychological models that capture fundamental human needs&Age group, Gender, Degree of Neuroticism in the Big Five personality traits, Degree of Extraversion in the Big Five personality traits, Degree of Openness in the Big Five personality traits, Degree of Agreeableness in the Big Five personality traits, Degree of Conscientiousness in the Big Five personality traits, Preference for a certain food, Preference for a certain living environment, Sleep preference, Investment preference, Saving preference, Importance placed on physical safety, Importance placed on environmental safety, Preference for the depth of interaction, Approach to handling conflict, Communication style in social settings: more concise or detailed, Need for a certain work environment, Need for recognition from others, Need for personal achievement, Preference for a certain knowledge area, Preference for a certain learning style, Preference for a certain form of creative expression (such as art, writing, music), Need for Order (neatness, organization, avoiding chaos), Need for Retention (to keep possession over an object, unwilling to lose or change), Need for Inviolacy (maintaining dignity and reputation, unviolated or undamaged), Need for Infavoidance (avoiding failure and embarrassment), Need for Counteraction (attempting to compensate for failure by trying again, with a desire to overcome obstacles and seek pride), Need for Seclusion (desire for isolation from others, maintaining privacy), Need for Dominance (controlling the environment or others through command or persuasion), Need for Deference (worship or obedience to others, following authority, admiration, or rules), Need for Autonomy (resisting influence from others, pursuing independence and self-reliance), Need for Contrariance (pursuing uniqueness, being different, or opposing the norm), Need for Abasement (submittance and obedience to others, accepting blame and punishment, even enjoying pain or misfortune to some extent), Need for Aggression (controlling, punishing, or harming others through forceful means), Need for Affiliation (desire for close, loyal relationships, pleasing others, winning friendships and attention), Need for Rejection (isolating, excluding, or discarding oneself from negatively evaluated objects or people), Need for Nurturance (assisting the weak, caring for others, protecting them from danger), Need for Succorance (desire to have one's own needs met by others, including being loved, cared for, helped, forgiven, and comforted), Need for Play (enjoying fun, relaxation, laughter)\\
    \midrule
Contemporary research in recommender systems and AI alignment that reflect social-cognitive needs in digital interactions & Degree of concern for the harmlessness of the statement, Degree of concern for the instruction-following of the statement, Degree of concern for the honesty of the statement, Degree of concern for the truthfulness of the statement, Degree of concern for the helpfulness of the statement, Degree of concern for the coherence of the statement, Degree of concern for the complexity of the statement\\
    \midrule
Content platform indicators that represent everyday
user needs & science, knowledge, psychology, cinema, entertainment, gaming, parenting, wild imagination, anime, sports, law, workplace, pets, travel, health, stories, cars, gourmet food, education, current events, home decor, international, finance, campus life, digital technology, emotions, humor, music, reading, painting, dance, crafts, photography, culture, fitness, art, stationery and planners, celebrities, outdoors, camping, social sciences, weddings, fashion\\
    \bottomrule
    \end{tabular}
    \end{adjustbox}
    \label{tab:space}
\end{table}

\setcounter{AlgoLine}{0}  
\begin{algorithm}
\small
\caption{Constructing the \textsc{AlignX} Dataset}\label{algorithm1}
\KwData{$x$: A user post; $Y=\{y_1,y_2,\cdots,y_N\}$: $N$ responses to the post; $\mathcal{M}=\{(x_m, Y_{m})\}_{m=1}^M$: The raw data; $\mathcal{D}=\{\mathcal{D}_1,\mathcal{D}_2,\cdots,\mathcal{D}_D\}:$ Preference dimensions in the preference space; $t$: Threshold for measuring intensity embedding similarity; $H$: Number of examples in behavioral personas} 
\KwResult{$\mathcal{P}_{\textsc{ugc}}$: User-Generated Content; $\mathcal{P}_{\textsc{pair}}$: Pair-wise comparative feedback; $\mathcal{P}_{\textsc{demo}}$: Demographic information; $y_w$: The preferred response; $y_l$: The less preferred response.}

\tcpor{Constructing Preference Pairs (\S\ref{pair_construct})}
\tcpo{(1) Intensity Annotation}
\While{$n \leftarrow 1$ \KwTo $N$}{
    \While{$d \leftarrow 1$ \KwTo $D$}
    {$l_d^n=\text{Annotation}(x, y_n, \mathcal{D}_d)$ \tcp{\textsf{Prompt \#1 in \S\ref{prompts}; $L$, the number of possible intensity levels, is set to 3}}}
}\par
\tcpo{(2) Intensity-based Clustering}
$\boldsymbol{l}^n=\text{Concatenate}(\{\text{one\_hot}(l_d^n)\}_{d=1}^D)\in\mathbb{R}^{DL}$ \par
$\mathcal{C}_1,\mathcal{C}_2,\cdots,\mathcal{C}_K=\text{K-Means}(\{\boldsymbol{l}_n\}_{n=1}^N)$\tcp{\textsf{$K$, the number of clusters, is set to 3}}\par
\tcpo{(3) Pair Selection}
$\mathcal{C}_w,\mathcal{C}_l=\text{UniformSample}(\mathcal{C}_1,\mathcal{C}_2,\cdots,\mathcal{C}_K)$\par
$y_w=\text{UniformSample}(\mathcal{C}_w)$\par
$y_l=\text{UniformSample}(\mathcal{C}_l)$\par\;

\tcpor{Constructing Personas (\S\ref{explicit_construct})}\par
\tcpo{Constructing $\mathcal{P}_{\textsc{demo}}$}\par
$\mathcal{P}_{\textsc{demo}}=\text{Annotation}(y_w,y_l, P)$ \tcp{\textsf{Prompt \#2 in \S\ref{prompts}}}\par\;

\tcpo{Constructing $\mathcal{P}_{\textsc{ugc}}$}\par
$\mathcal{P}_{\textsc{ugc}}=\{~\}$\par
\While{True}{
    ($x', Y')=\text{UniformSample}(\mathcal{M}\setminus\{(x,Y)\}))$\par
        $y' = \text{argmax}_{y'\in Y'} \text{Similarity}(y', y_w)$ \tcp{\textsf{cosine similarity between their intensity embeddings}}\par
        \If{the similarity score between $y'$ and $y_w$ is larger than $t$=0.6}{
            $\mathcal{P}_{\textsc{ugc}}\text{.add}((x',y'))$
        }
    }
Retaining $H$ examples in $\mathcal{P}_{\textsc{ugc}}$, making it satisfy the preference consistency criterion.
\par\;

\tcpo{Constructing $\mathcal{P}_{\textsc{pair}}$}\;
$\mathcal{P}_{\textsc{pair}}=\{~\}$\par
\While{True}{
    Sampling $(x, y_w',y_l')$ from all preference pairs constructed in \S\ref{pair_construct}\par
        \If{the similarity scores between $y_w'$ and $y_w$, as well as between $y_l'$ and $y_l$ are larger than $t$=0.6 }{
            $\mathcal{P}_{\textsc{pair}}\text{.add}((x',y_w', y_l'))$
        }
    }
Retaining $H$ examples in $\mathcal{P}_{\textsc{pair}}$, making it satisfy the preference consistency criterion.

{\Return $\mathcal{P}_{\textsc{ugc}}, \mathcal{P}_{\textsc{pair}}, \mathcal{P}_{\textsc{demo}}, y_w, y_l$}
\end{algorithm}

\subsection{Data construction pipeline}\label{app:cons_pipeline}
Alg.~\ref{algorithm1} describes the construction pipeline for constructing the \textsc{AlignX} data from any raw data providing multiple responses $Y=\{y_1,y_2,\cdots,y_N\}$ to the same post $x$. We use Qwen2.5-14B-Instruct for intensity annotation and demographic information annotation.

\subsection{Annotation prompts}\label{prompts}

\begin{tcolorbox}[colframe=pink!90!black, colback=pink!10!white, coltitle=black, fonttitle=\bfseries, title=Intensity Annotation Prompt, breakable]

\textbf{Example 1:}\\

\textbf{Query:}\\
I’m at urgent care because I fucked up my ankle. The conversation went something like this.\\
Nurse: Any surgeries in the past?\\
Me: Yes, a bilateral salpingectomy in September of last year.\\
And, as she’s about to leave: Any chance you’re pregnant?\\
Me: No\\
Me (thinking): she has no idea what a bilateral salpingectomy is. Kind of funny at urgent care just now\\

\textbf{Comment:}\\
Dr.: any chance you are pregnant?\\
Me: only with a turd, anticipated birthday tomorrow.\\

\textbf{Task:}\\
What does this comment suggest about the commenter's Neuroticism in the Big Five personality traits? (1) High Neuroticism (2) Low Neuroticism (3) Comments alone cannot reflect\\

\textbf{Response:}\\
2\\

\textbf{Task:}\\
What does this comment suggest about the commenter's Extraversion in the Big Five personality traits? (1) High Extraversion (2) Low Extraversion (3) Comments alone cannot reflect\\

\textbf{Response:}\\
1\\

\textbf{Task:}\\
Does this comment indicate that the commenter likes or dislikes ``painting''? (1) Likes (2) Dislikes (3) Comments alone cannot reflect\\

\textbf{Response:}\\
3\\

\textbf{Your task:}\\

\textbf{Query:}\\
\{Post\}\\

\textbf{Comment:}\\
\{Comment\}\\

\textbf{Task:}\\
\{Question\}\\

\textbf{Response:}\\
\end{tcolorbox}

\begin{tcolorbox}[colframe=pink!90!black, colback=pink!10!white, coltitle=black, fonttitle=\bfseries, title=Demographic Information Generation and Preference Direction Annotation Prompt, breakable]

Who is likely to prefer $<$Response 1$>$ and dislike $<$Response 2$>$? Compare the preferred and disliked responses, and describe a person using selected dimensions from the following list of 48, reflecting what can be observed:
\begin{enumerate}
    \item Age group: (``Young'', ``Older'')
    \item Gender: (``Female'', ``Male'')
    \item Occupation
    \item Degree of Neuroticism in the Big Five personality traits: (``High'', ``Low'')
    \item Degree of Extraversion in the Big Five personality traits: (``High'', ``Low'')
    \item Degree of Openness in the Big Five personality traits: (``High'', ``Low'')
    \item Degree of Agreeableness in the Big Five personality traits: (``High'', ``Low'')
    \item Degree of Conscientiousness in the Big Five personality traits: (``High'', ``Low'')
    \item Preference for a certain food: (``Likes'', ``Dislikes'')
    \item Preference for a certain living environment: (``Likes'', ``Dislikes'')
    \item Sleep preference: (``Likes'', ``Dislikes'')
    \item Investment preference: (``Aggressive'', ``Conservative'')
    \item Saving preference: (``Good at saving'', ``Bad at saving'')
    \item Importance placed on physical safety: (``Concerned'', ``Not concerned'')
    \item Importance placed on environmental safety: (``Concerned'', ``Not concerned'')
    \item Preference for the depth of interaction: (``Superficial interaction (casual, stress-free chat)'', ``Deep interaction (discussing interests, emotional topics, etc.)'')
    \item Approach to handling conflict: (``Direct communication'', ``Avoidance, mediation, compromise'')
    \item Communication style in social settings: more concise or detailed: (``Concise'', ``Detailed'')
    \item Need for a certain work environment: (``Strong'', ``Mild'')
    \item Need for recognition from others: (``Strong'', ``Mild'')
    \item Need for personal achievement: (``Strong'', ``Mild'')
    \item Preference for a certain knowledge area: (``Likes'', ``Dislikes'')
    \item Preference for a certain learning style: (``Likes'', ``Dislikes'')
    \item Preference for a certain form of creative expression (such as art, writing, music): (``Likes'', ``Dislikes'')
    \item Need for Order (neatness, organization, avoiding chaos): (``Strong'', ``Indifferent'')
    \item Need for Retention (to keep possession over an object, unwilling to lose or change): (``Strong'', ``Indifferent'')
    \item Need for Inviolacy (maintaining dignity and reputation, unviolated or undamaged): (``Strong'', ``Indifferent'')
    \item Need for Infavoidance (avoiding failure and embarrassment): (``Strong'', ``Indifferent'')
    \item Need for Counteraction (attempting to compensate for failure by trying again, with a desire to overcome obstacles and seek pride): (``Strong'', ``Indifferent'')
    \item Need for Seclusion (desire for isolation from others, maintaining privacy): (``Strong'', ``Indifferent'')
    \item Need for Dominance (controlling the environment or others through command or persuasion): (``Strong'', ``Indifferent'')
    \item Need for Deference (worship or obedience to others, following authority, admiration, or rules): (``Strong'', ``Indifferent'')
    \item Need for Autonomy (resisting influence from others, pursuing independence and self-reliance): (``Strong'', ``Indifferent'')
    \item Need for Contrariance (pursuing uniqueness, being different, or opposing the norm): (``Strong'', ``Indifferent'')
    \item Need for Abasement (submittance and obedience to others, accepting blame and punishment, even enjoying pain or misfortune to some extent): (``Strong'', ``Indifferent'')
    \item Need for Aggression (controlling, punishing, or harming others through forceful means): (``Strong'', ``Indifferent'')
    \item Need for Affiliation (desire for close, loyal relationships, pleasing others, winning friendships and attention): (``Strong'', ``Indifferent'')
    \item Need for Rejection (isolating, excluding, or discarding oneself from negatively evaluated objects or people): (``Strong'', ``Indifferent'')
    \item Need for Nurturance (assisting the weak, caring for others, protecting them from danger): (``Strong'', ``Indifferent'')
    \item Need for Succorance (desire to have one's own needs met by others, including being loved, cared for, helped, forgiven, and comforted): (``Strong'', ``Indifferent'')
    \item Need for Play (enjoying fun, relaxation, laughter): (``Strong'', ``Indifferent'')
    \item Degree of concern for the harmlessness of the statement: (``Strong'', ``Indifferent'')
    \item Degree of concern for the instruction-following of the statement: (``Strong'', ``Indifferent'')
    \item Degree of concern for the honesty of the statement: (``Strong'', ``Indifferent'')
    \item Degree of concern for the truthfulness of the statement: (``Strong'', ``Indifferent'')
    \item Degree of concern for the helpfulness of the statement: (``Strong'', ``Indifferent'')
    \item Degree of concern for the coherence of the statement: (``Strong'', ``Indifferent'')
    \item Degree of concern for the complexity of the statement: (``Strong'', ``Indifferent'')
\end{enumerate}

Compare the liked and disliked responses, and derive the keywords for likes and dislikes, ensuring the keywords are selected from the following list:\\

\{science, knowledge, psychology, cinema, entertainment, gaming, parenting, wild imagination, anime, sports, law, workplace, pets, travel, health, stories, cars, gourmet food, education, current events, home decor, international, finance, campus life, digital technology, emotions, humor, music, reading, painting, dance, crafts, photography, culture, fitness, art, stationery and planners, celebrities, outdoors, camping, social sciences, weddings, fashion\}.\\

If keywords cannot be summarized, please leave them blank.\\

\textbf{Example 1:}\\

\textbf{Query:}\\
Sure, kids are expensive and a lot of people who would otherwise want them are definitely putting off parenthood because they can't afford it. But everyone acts like this is the only reason why younger adults don't want kids, and it's still not okay to admit that you just don't want to be a parent regardless of money.\\

I’m sick of mentioning my student debt and getting sympathy from people who feel bad that I can’t afford to be a mom, when I never wanted to be one to begin with. Nobody can handle the truth about how a lot of millennial women aren’t having kids because for the first fucking time in history, we actually have the conscious choice not to, and have access to birth control that actually works.\\

I’m sick of people putting all the blame on finances in articles about declining birth rates and childfree millennials\\

\textbf{Output:}

$<$Response 1$>$\\
I don't quite see how we're the problem, when we're the ones who are trying to work, get a decent job in our field, maybe move out of our parents house like a normal, functional human being, rather than popping out kids left and right that we cannot afford as soon as we finish high school.\\

$<$Response 2$>$\\
Women with or without student loan debt have many of the same problems. The wildly (IMO) outrageous childcare costs and depending on the area high housing costs are hard for two incomes and impossible for one. Young women are encouraged to have children and yet its been proven, income and promotion opportunities will be stalled or decreased. If you add student loan debt it just becomes more untenable.\\

Full disclosure, I'm a childfree, 65 year-old. Sadly, we had the same issues with sexual harassment and lack of advancement, but we did not have anywhere near the financial burdens. Unless you wanted to go to an Ivy League college you could work your way through college with part-time jobs or a little help from your parents and get a well-paying job. Housing was reasonably priced as was childcare.\\

Old people, sadly I have to include myself, need to get a grip and understand how much times have changed and raising and educating a child really costs. Maybe when you all run the government it will change.\\
\\
\{
    ``reason'': ``People who prefer $<$Response 1$>$ value personal freedom and see not having children as an individual choice, not just a result of financial or societal pressures. They reject traditional views on parenthood and feel $<$Response 2$>$'s focus on financial burdens limits their freedom to choose.'', 
    
    ``textural persona'': ``A well-educated, ambitious founder of a tech startup who is willing to challenge traditional views on parenting and society. He takes a cautious approach to having children, valuing conscious family planning choices and seeing economic factors as secondary to the broader societal shift in expectations. He is unlikely to conform to societal norms, focusing on personal goals and independence, and takes responsibility for their financial and career decisions.'', 
    
    ``dimension description'': \{
        ``Age group'': ``Young'', 
        ``Occupation'': ``Founder of a tech startup'', 
        ``Degree of Openness in the Big Five personality traits'': ``High'', 
        ``Degree of Agreeableness in the Big Five personality traits'': ``Low'', 
        ``Degree of Conscientiousness in the Big Five personality traits'': ``High'', 
        ``Preference for a certain living environment'': ``Likes'', 
        ``Approach to handling conflict'': ``Direct communication'', 
        ``Need for a certain work environment'': ``Strong'', 
        ``Need for personal achievement'': ``Strong'', 
        ``Need for Autonomy (resisting influence from others, pursuing independence and self-reliance)'': ``Strong''
    \}, 
    
    ``like keywords'': ``education, workplace, finance'', 
    
    ``dislike keywords'': ``parenting''
\}\\

\textbf{Your task:}\\

\textbf{Query:}\\
\{Post\}\\

\textbf{Output:}\\
$<$Response 1$>$\\
\{Response 1\}\\

$<$Response 2$>$\\
\{Response 2\}\\

Now, please output the persona you created, keywords, and a short rationale below in a JSON format by filling in the placeholders in []:\\
\{``reason'': ``[your rationale]'', ``textural persona'': ``[persona you created]'', ``dimension description'':``[description of corresponding dimensions]'', ``like keywords'': ``[like, Optional]'', ``dislike keywords'': ``[dislike, Optional]''\}

\end{tcolorbox}

\begin{tcolorbox}[colframe=pink!90!black, colback=pink!10!white, coltitle=black, fonttitle=\bfseries, title=GPT-4 Judgment Prompt, breakable]

The given user behavior history reflects the user's preferences. For a new query, which response better matches the user's needs and preferences, allowing for a tie between the two responses?\\

\textbf{User Behavior History}\\
\{Concatenated user history information\}\\

\textbf{Query}\\
\{Query\}\\

\textbf{Response A}\\
\{Response A\}\\

\textbf{Response B}\\
\{Response B\}\\

Now, please output your choice below in a JSON format by filling in the placeholders in []:

\{``choice'': ``[Response A/Response B/Tie]''\}

\end{tcolorbox}




\section{Alignment methods}
\subsection{Conversion of preference direction to natural language description}\label{conversion}
When a persona $\mathcal{P}$ contains multiple components (e.g., both $\mathcal{P}_{\textsc{ugc}}$ and $\mathcal{P}_{\textsc{demo}}$) or behavioral personas comprise multiple instances, we first aggregate their preference directions into a unified vector $\Tilde{P} \in \mathbb{R}^d$, where $d$ denotes the dimension of preference space. Specifically, for each preference dimension $i$, we convert the categorical preferences across all components into numerical values (positive → 1.0, neutral → 0.5, negative → 0.0) and compute their average. We then determine the direction for each dimension based on two thresholds: dimensions with average values above $t_1$ are assigned positive direction, those below $t_2$ are assigned negative direction, and the remaining are considered neutral, resulting in the final preference direction vector $\Tilde{P}$.

We then map $\Tilde{P}$ to natural language descriptions following systematic linguistic rules. For dimensions with positive direction, we prepend ``High'' to personality traits (e.g., ``High Neuroticism''), ``Prefers'' to communication styles (e.g., ``Prefers detailed communication''), and use positive forms for values (e.g., ``Values privacy''). For dimensions with negative direction, we prepend ``Low'' to personality traits (e.g., ``Low Extraversion''), ``Avoids'' to communication styles (e.g., ``Avoids confrontational communication''), and use negative forms for values (e.g., ``Disregards material wealth'').

The final description is constructed by concatenating all non-neutral preference descriptions with comma separators. This systematic conversion process ensures consistent and interpretable preference articulation while preserving the key signals from the original vector.

\subsection{Training examples}
\label{app: training_examples}
In this section, we present the prompt format used during the training of ICA and PBA.

\begin{tcolorbox}[colframe=pink!90!black, colback=pink!10!white, coltitle=black, fonttitle=\bfseries, title=Data Format for ICA. Case:, breakable]

Generate a task-specific response based on user historical behavior and preferences.\\

\textbf{Task}

My 10-year-old daughter has ADHD, Dyslexia, Dysgraphia, and Dyscalculia, with an IQ of 79. She struggles academically and socially but loves theatre. Looking for advice or similar experiences. Thank you.\\

\textbf{Historical Behavior and User Preferences}\\

\textbf{This person has commented on some posts:}
\begin{enumerate}
    \item \textbf{Post:}
    I’ve been trying to potty train my daughter, but today I let her go without a diaper, and she stayed dry all day. She’s trained, just not potty trained! \\
    \textbf{Comment:}
    Oh, so she’s ``trained''—just not in the way that actually matters. Maybe try using the potty next time instead of just hoping for magic!
\end{enumerate}

\textbf{This person has chosen or rejected comments on some posts:}
\begin{enumerate}
    \item \textbf{Post:}
    I'm potty training a stubborn boy and want to use cheap, uncomfortable diapers or training pants to make him prefer big boy pants. Any recommendations for the worst ones?\\
    \textbf{Chosen:}
    Put him in a swim diaper—it will leak as soon as he urinates, making him uncomfortable. Tell him he's outgrown pull-ups, so he needs to use the toilet and wear big boy pants.\\
    \textbf{Rejected:}
    I once read about a dad who put his kid in a smaller diaper, acted surprised, and said, "Oh no! You've outgrown diapers! Time for big boy underwear." It worked for him.
\end{enumerate}

\textbf{This person's persona is:}\\
A young executive with low agreeableness prefers direct communication in social settings and values independence and efficiency. This person likes parenting, education, dislikes emotions.\\

\textbf{Response:}

\end{tcolorbox}

\begin{tcolorbox}[colframe=pink!90!black, colback=pink!10!white, coltitle=black, fonttitle=\bfseries, title=Data Format for PBA. Case:, breakable]

Generate a task-specific response based on user preferences.\\

\textbf{Task}

My 10-year-old daughter has ADHD, Dyslexia, Dysgraphia, and Dyscalculia, with an IQ of 79. She struggles academically and socially but loves theatre. Looking for advice or similar experiences. Thank you.\\

\textbf{User Preferences}

Low Agreeableness, Prefers direct communication to handle conflict, Concise communication style, Strong need for Autonomy (pursuing independence and self-reliance), Likes parenting, Likes education, Dislikes emotions\\

\textbf{Response:}

\end{tcolorbox}

\section{Details of the experimental setup}
\label{appendix:setup}

\subsection{Benchmarks}
\label{benchmarks_detail}

We conduct experiments on four benchmarks: (1) UF-P-4~\citep{poddar2024personalizing}: An alignment dataset focused on universal values (e.g., helpfulness) with consistently positive preference directions. We treat each preference pair as a test case and construct $\mathcal{P}_{{\textsc{pair}}}$ by randomly sampling 2$\sim$8 pairs from the same dimension. (2) PRISM~\citep{kirk2024prism}: A collection of real user-LLM interaction preferences with unknown preference dimensions and directions.\footnote{Manual inspection finds that most preference pairs in PRISM often reflect various combinations of universal values (e.g., more friendly, more empathy).} Each preference pair serves as a test case, with $\mathcal{P}_{{\textsc{pair}}}$ comprising all other pairs from the same user. (3) P-\textsc{Soups}~\citep{jang2023personalized}: A benchmark spanning three novel dimensions beyond our preference space, with bidirectional preference. For each dimension-direction configuration, we use individual preference pairs as test cases and sample 4 pairs from the remaining as $\mathcal{P}_{{\textsc{pair}}}$. (4) \textsc{AlignX}$_{\rm test}$: A comprehensive test set constructed following the methodology in \S\ref{explicit_construct}, covering all dimensions in our preference space with balanced positive and negative directions. For each dimension and direction combination, we construct dedicated test cases where both the preference pair and persona consistently exhibit the specified direction for the given dimension. Our evaluation is both on individual persona types and arbitrary combinations of three types of personas in $\mathcal{P}$ to test the efficacy under varying levels of persona completeness. Both $\mathcal{P}_{\textsc{ugc}}$ and $\mathcal{P}_{\textsc{pair}}$ in \textsc{AlignX}$_{\rm test}$ include 4 randomly sampled examples. 
We ensure that none of the above test cases overlap with training data.

\subsection{Implementation details}\label{implementation_details}
We implement ICA and PBA based on Llama-3.1-8B-Instruct. Both models are trained on 8 A100 GPUs for one epoch using Adam optimizer \citep{kingma2014adam}, DeepSpeed with ZeRO-3 \citep{rajbhandari2020zero} and Flash-attention-2 \citep{dao2023flashattention}, with the following configuration: learning rate of 5e-7, batch size of 128, maximum sequence length of 8,192, and $\beta=0.1$ for both Eq.~\ref{dpo_ica} and~\ref{dpo_pba}. During training, both $\mathcal{P}_{\textsc{ugc}}$ and $\mathcal{P}_{\textsc{pair}}$ retain only 0$\sim$4 randomly sampled examples for computational efficiency. We use Qwen2.5-14B-Instruct for annotating preference directions in PBA. \S\ref{app: training_examples} describes the format used during the training of ICA and PBA.

\section{Supplementary experimental results}
\label{app_res}

\subsection{Detailed scores on the P-\textsc{Soups} benchmark}\label{app_psoup}

The P-\textsc{Soups} benchmark involves three preference dimensions: \textit{Expertise}, \textit{Informativeness}, and \textit{Style}. Although these dimensions are not 
aincluded in the dimensions defined by \textsc{AlignX}, they can be represented as combinations of \textsc{AlignX}'s preference space. The experimental results in Table \ref{tab:psoup} show that \textsc{AlignXpert} (trained on the 7\% subset of \textsc{AlignX} data) significantly outperforms the baselines in both individual and combined dimensions.

\begin{table}[!th]
  \caption{Accuracy performance (\%) of each dimension on P-\textsc{Soups}.}
  \label{tab:psoup}
  \centering
  \begin{tabular}{@{}lccc@{}}
    \toprule
    \textbf{Model} & \textbf{Expertise} & \textbf{Informativeness} & \textbf{Style}\\
    \midrule
    \textbf{Llama-3-8B-Instruct} & 40.17 & 56.81 & 51.33\\
    \textbf{Qwen2.5-7B-Instruct} & 38.00 & 43.69 &22.00\\
    \textbf{Mistral-7B-Instruct-v0.2} & 40.17 & 46.35 & 33.83\\
    \midrule
\textbf{\textsc{AlignXpert}$_{\textsc{ica}}$} & \textbf{85.17} & \underline{60.80} & \textbf{83.67}\\
\textbf{\textsc{AlignXpert}$_{\textsc{pba}}$} & \underline{84.50} & \textbf{66.11} & \underline{79.17}\\
    \bottomrule
  \end{tabular}
\end{table}

\subsection{Results with larger models as references}\label{app_larger}

To validate the effectiveness of AlignXpert, we use Llama-3-70B-Instruct as the reference model and conduct experiments on both in-domain (AlignX$_{\rm test}$) and out-of-domain (P-Soups) benchmarks. Table \ref{tab:larger} shows that our method demonstrates strong performance compared to both the 70B large model and the baselines.

\begin{table}[!th]
  \caption{Alignment accuracy (\%) of different models using Llama-3-70B-Instruct as the reference model.}
  \label{tab:larger}
  \centering
  \begin{tabular}{lcm{0.001em}ccccc}
    \toprule
    \multirow{2}{*}{\textbf{Model}} & \multirow{1}{*}{\textbf{P-\textsc{Soups}}} && \multicolumn{4}{c}{\textbf{\textsc{AlignX}$_{\rm test}$}}\\
    \cmidrule{2-2}
    \cmidrule{4-8}
    &$\mathcal{P}_{{\textsc{pair}}}$ && $\mathcal{P}_{{\textsc{ugc}}}$ & $\mathcal{P}_{{\textsc{pair}}}$ & $\mathcal{P}_{\textsc{demo}}$ & $\mathcal{P}$ & \\
    \midrule
    \textbf{Llama-3-8B-Instruct} & 46.78 && 50.59 & 51.32 & 47.85 & 50.69\\
    \textbf{Qwen2.5-7B-Instruct} & 34.62 && 48.98 & 49.00 & 50.67 & 49.87 \\
    \textbf{Mistral-7B-Instruct-v0.2} & 38.95 && 49.38 & 47.95 & 48.17 & 48.27 \\
    \midrule
    \multicolumn{8}{l}{\textit{Training with Full Dataset of 1,311,622 Samples (100\%)}}\\
    \textbf{\textsc{AlignXpert}}$_{\textsc{ica}}$ & 59.39  && \textbf{57.80} & 57.64 & \textbf{90.93} & \textbf{73.21} \\
    \textbf{\textsc{AlignXpert}}$_{\textsc{pba}}$ & \textbf{80.20} && 56.70 & \textbf{59.66} & 88.19 & 71.01\\
    \bottomrule
  \end{tabular}
\end{table}

\section{Analysis}
\label{appendix:analysis}
\subsection{Preference diversity}
We analyze potential correlations between the 90 dimensions across all 1.3 million examples. The Pearson correlation heatmap is available at Figure \ref{fig:heatmap}, with dimension indices correspond to Table \ref{tab:space}. Among all dimension pairs, 99.65\% show correlations below 0.5, indicating that our 90 dimensions capture largely independent aspects of user preferences. Only two dimensions—``Need for Abasement (submission and obedience to others, accepting blame and punishment, even deriving pleasure from pain or misfortune)'' and ``Need for Aggression (controlling, punishing, or harming others through force)''—exhibit a strong correlation (0.86). These two may reflect different expressions of the same underlying personality trait and can complement each other in practice without affecting other dimensions. Overall, at least 89 independent dimensions comprehensively capture user personality.

Based on the high independence among the 90 dimensions, we calculate that there are 1,210,986 different preference directions in \textsc{AlignX}, accounting for 92.33\% of the total data, with 1,210,837 preference directions coming from Reddit annotated data, demonstrating diverse user preferences.

\subsection{Annotation bias}
\label{Annotation_bias}
To construct data at scale, we use LLM for annotation during the process. We take several measures to assess potential LLM annotation biases.

In the qualitative analysis, our process first clusters responses into three groups and randomly selects responses from different groups as $y_w$ and $y_l$ to determine preference direction (\S\ref{pair_construct}). LLMs only label these pre-determined preferences without generating opinions, thus minimizing bias introduction.

In the quantitative analysis, we randomly select 100 samples from the data labeled by Qwen2.5-14B-Instruct in \textsc{AlignX}, and replace the annotation model with GPT-4 and recruit 4 crowdsource annotators to relabel these samples, verifying the bias introduced by LLM selection and whether LLMs themselves introduce bias. Additionally, we use a case study to analyze whether there exists societal bias in the data.

\paragraph{Bias from LLM selection}
We re-label the intensity of 100 preference pairs across different dimensions using gpt-4-turbo-2024-04-09. We calculate that 84.78\% of the obtained preference directions match those labeled by Qwen2.5-14B-Instruct, indicating that our method demonstrates robustness in selecting LLMs.

\paragraph{Bias from LLM annotation}
We engage four crowdsource annotators for the labeling task. Human evaluation of 100 random samples from four crowdsource annotators shows 84.15\% complete agreement with LLM annotations across all dimensions (while different people's annotations show 90.83\% consistency), validating annotation reliability.

These annotators demonstrate excellent English proficiency, allowing them to accurately assess preference tendencies. Regarding human annotations, the process is reviewed and approved by relevant institutions. After a fair evaluation of the workload, we compensate each evaluator \$5.60 per 10 samples.

\paragraph{Case study}
We identify two attributes often linked to societal bias: ``Gender'' and ``Need for Succorance (desire to have one's own needs met by others, including being loved, cared for, helped, forgiven, and comforted).'' Table \ref{tab:case} shows the data count for their different combinations.

\begin{table}[]
    \centering    
    \caption{The data count for different combinations of ``Gender'' and ``Succorance'' at varying intensities.}
    \resizebox{0.43\linewidth}{!}{
    \begin{tabular}{lcc}
    \toprule
     \textbf{Gender/Succorance}& \textbf{Negative} & \textbf{Positive}  \\
     \midrule
     \textbf{Negative} & 1,986 & 1,980 \\
     \textbf{Positive} & 2,117 & 1,973 \\
     \bottomrule
    \end{tabular}}
    \label{tab:case}
\end{table}

The data distribution is relatively balanced without clear tendencies, indicating that our data does not introduce bias.

\subsection{Complexity of preference dimensions}
To investigate the complexity of the 90 dimensions, we analyze the computational challenges encountered during practical deployment. Only \textsc{AlignXpert}$_{\textsc{pba}}$ involves the 90-dimensional space in this process. The main computational aspects are as follows:

$\bullet$
Predicting preference directions from Persona $\mathcal{P}$: With 8 examples in $\mathcal{P}$, computation time increases modestly from 10.20s to 16.21s as dimensions increase from 1 to 90 - a reasonable overhead given the benefits of comprehensive preference modeling.

$\bullet$
Converting preference directions to natural language: This rule-based transformation has negligible computational cost.

Overall, the 90 dimensions provide more comprehensive preference coverage compared to existing research while maintaining manageable computational overhead.

\section{Impact statement} 
\label{Impact}
This work aims to advance personalized AI alignment, which carries both promising benefits and potential risks that warrant careful consideration. On the positive side, our framework could lead to AI systems that better serve diverse user groups, potentially reducing systematic biases against underrepresented populations and improving user satisfaction across different cultural backgrounds. The ability to adapt to individual preferences could make AI assistance more accessible and effective for users with varying needs, education levels, and communication styles.

However, several ethical considerations must be addressed. First, while personalization can enhance user experience, it may also reinforce existing biases or create echo chambers if not properly constrained. Second, the collection and use of personal preference data raises privacy concerns, requiring careful attention to data protection and user consent. Third, there's a need to balance individual preferences with broader societal values and ethical principles, ensuring that personalization doesn't compromise fundamental moral standards.

To mitigate these risks, we recommend: (1) implementing robust privacy protection mechanisms for persona data, (2) developing clear guidelines for balancing personal preferences with ethical constraints, and (3) regularly auditing personalized models for potential bias amplification. We believe that responsible development of personalized AI alignment, with appropriate safeguards, can contribute positively to making AI systems more inclusive and beneficial for all users.

\section{Limitations}
\label{Limitations}
Due to the difficulty in obtaining large-scale real user-LLM data, even though we incorporate datasets such as PKU-SafeRLHF, UltraFeedback, and HelpSteer2 to cover LLM-specific scenarios, interactions from Reddit still represent a significant portion of AlignX. As a result, AlignX does not contain a substantial amount of real-world data and serves primarily as a testbed for scaling personalized preferences for user-level alignment. To address this limitation, when human-LLM interaction data (e.g., dialogues, user preferences) become available in the future, our platform-agnostic methodology can be directly applied to obtain alignment data.

\end{document}